\documentclass[journal,twoside,print]{IEEEtran}

\usepackage{cite}
\usepackage{amsmath,amssymb,amsfonts}
\usepackage{graphicx}
\usepackage{algorithm}
\usepackage{algorithmic} %

\usepackage{booktabs}
\usepackage{array}
\usepackage{multirow}
\usepackage[table]{xcolor}
\definecolor{mygray}{gray}{0.9}

\usepackage[caption=false,font=footnotesize]{subfig}

\usepackage{url}

\usepackage[hidelinks]{hyperref}

\definecolor{subsectioncolor}{rgb}{0,0,0}

 \usepackage{textcomp}
 \makeatletter
\def\ps@titlepagestyle{%
  \def\@oddhead{%
    \begin{tabular}{@{}p{\textwidth}@{}}\\[-32pt]%
      {\vbox{%
        \hsize\textwidth
        \scriptsize\textsf{PREPRINT \hfill \thepage}\par
        \vspace*{-3pt}%
        \vbox{\color{subsectioncolor}\hrule height1pt width\textwidth depth0pt}%
      }}%
    \end{tabular}%
  }%
  \def\@evenhead{\@oddhead}%
  \def\@oddfoot{}%
  \def\@evenfoot{}%
}
\makeatother
\begin{document}
\title{Co-Evolution-Based Metal-Binding Residue Prediction with Graph Neural Networks}
\author{Sayedmohammadreza Rastegari, Sina Tabakhi,~\IEEEmembership{Member,~IEEE}, Xianyuan Liu, Tianyi Jiang, Wei Sang, and Haiping Lu,~\IEEEmembership{Senior Member,~IEEE}%
\thanks{}%
\thanks{Sayedmohammadreza Rastegari is with the Faculty of Computer Engineering, University of Isfahan, Isfahan, Iran
        {\tt\small \{s.mohammadreza.rastegari@gmail.com\}}.}%
\thanks{Sina Tabakhi and Xianyuan Liu are with the School of Computer Science and Centre for Machine Intelligence, University of Sheffield, Sheffield, United Kingdom
        {\tt\small \{\{stabakhi1, xianyuan.liu\} @sheffield.ac.uk\}}.}%
\thanks{Tianyi Jiang is with the Institute of Cyberspace Security, College of Information Engineering, Zhejiang University of Technology, Hangzhou, China, and the School of Computer Science and Centre for Machine Intelligence, University of Sheffield, Sheffield, United Kingdom
    {\tt\small \{josieyi0319@163.com\}}.}%
\thanks{Wei Sang is with the School of Basic Medical Sciences and Institute of Medical Technology, Shanxi Medical University, Taiyuan, China
    {\tt\small \{sangwei@sxmu.edu.cn\}}.}%
\thanks{Haiping Lu is with the School of Computer Science and Centre for Machine Intelligence, University of Sheffield, Sheffield, United Kingdom, and Institute of Big Data Science and Industry, Shanxi University, Taiyuan, China
        {\tt\small \{h.lu@sheffield.ac.uk\}}.}%
\thanks{Corresponding authors: Wei Sang; Haiping Lu.
}   
}

\maketitle

\begin{abstract}
Understanding protein-metal interactions is central to structural biology, with metal ions being vital for catalysis, stability, and signal transduction. Predicting metal-binding residues and metal types remains challenging due to the structural and evolutionary complexity of proteins.
Conventional sequence- and structure-based methods often fail to capture co-evolutionary constraints that reflect how residues evolve together to maintain metal-binding functionality. Recent co-evolution-based methods capture part of this information, but still underutilize the complete co-evolved residue network.
To address this limitation, we introduce the Metal-Binding Graph Neural Network (MBGNN), which leverages the complete co-evolved residue network to better capture complex dependencies within protein structures. 
The experimental results show that MBGNN substantially outperforms the state-of-the-art co-evolution-based method, MetalNet2, achieving F1 score improvements of 2.5\% for binding residue identification and 3.3\% for metal type classification on the MetalNet2 dataset. 
Its superiority is further demonstrated on both the MetalNet2 and MIonSite datasets, where it outperforms two co-evolution-based and two sequence-based methods, achieving the highest mean F1 scores across both prediction tasks.
These findings highlight how integrating co-evolutionary residue networks with graph-based learning advances our ability to decode protein-metal interactions, thereby facilitating functional annotation and rational metalloprotein design. 
The code and data are released at \url{https://github.com/SRastegari/MBGNN}.
\end{abstract}


\section{Introduction}

\begin{figure*}[t]
    \centering
    \includegraphics[width=\textwidth]{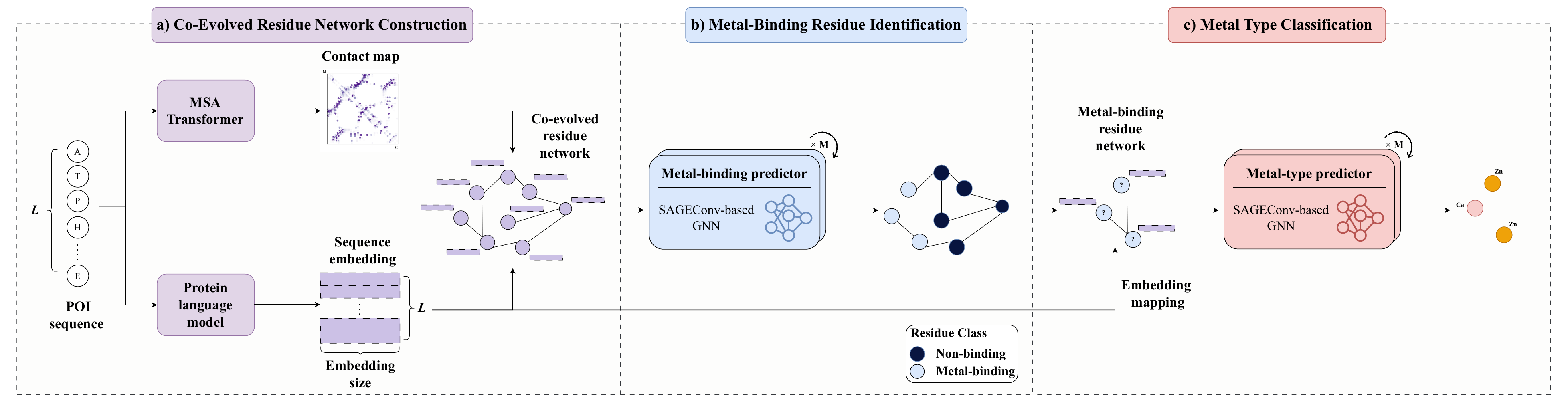}
    \caption{Overview of Metal-Binding Graph Neural Network (MBGNN) pipeline.
    (a) Co-evolved network construction begins with the chain sequence of a protein of interest (POI) and multiple sequence alignment (MSA). The MSA Transformer identifies co-evolved residue pairs, while a protein language model (PLM) generates residue-level embeddings. Identified residue pairs are assembled into co-evolved residue networks, with each residue represented by its corresponding PLM-derived embedding.
    (b) The metal-binding predictor processes these networks using graph neural networks (GNNs) trained under an \textit{M}-fold ensemble strategy. Predictions are obtained by averaging probabilities across models to identify metal-binding residues.
    (c) Predicted metal-binding residues are then assembled into new co-evolved networks, with residues again mapped to their PLM-derived embeddings. The metal-type predictor uses GNNs trained with the same \textit{M}-fold ensemble strategy to classify each network into 11 metal types.}
    \label{fig:visualabstract}
    \vspace{-5pt}
\end{figure*}

\IEEEPARstart{C}{haracterizing} protein interactions with small molecules and ions, known as ligands, provides mechanistic insight into enzymatic activity, conformational dynamics, and other determinants of protein function.
These interactions drive essential cellular processes \cite{wang2024multimodrlbp,yi2024equivariant,p1,liu2025interpretable,p2,p3,gladyshev2013comparative} and have important applications in drug discovery and biotechnology. Among various ligands, metal ions play a particularly significant role by binding to specific protein sites and contributing to structural stability and catalysis. More than one-third of proteins in the proteome can bind metal ions \cite{cheng2023co}, making it essential to identify metal-binding residues -- the specific amino acids that facilitate these interactions.
However, detecting these residues is challenging due to the complexity of protein structures, the hidden nature of binding sites, the diversity of metal ions, and the limited number of residues involved in these interactions. Furthermore, as experimental methods are costly and time-consuming, computational approaches offer faster and more scalable alternatives for accurate predictions.

There are three main computational approaches for predicting metal-binding sites~\cite{XIA2024102793, ye2022comprehensive}. \textit{Structure-based} methods directly analyze features from the three-dimensional coordinates of a protein~\cite{lin2016mib, yi2024equivariant, 10.1093/bioinformatics/btac534, durr2023metal3d}. Their main limitation is dependence on experimentally determined or accurately predicted structures, which are not always available. To address this, \textit{sequence-based} methods are developed to use only amino acid sequences~\cite{lmetalsite, mionic, cui2019predicting, 9772410} and they are scalable to broader applications. However, these methods struggle to capture the full complexity of binding interactions from sequences alone~\cite{10.1093/nar/gkab044}, and many implementations are trained for only a limited range of metal types~\cite{mionic}. \textit{Multimodal} methods have recently been introduced to integrate structural and sequence information, combining the strengths of both~\cite{10.1093/bioinformatics/btaa110,liu2025interpretable}. Nevertheless, their reliance on structural information means they remain constrained by the availability of reliable protein structures. As a result, sequence-based methods remain the most practical solution for large-scale metal-binding prediction.

To improve sequence-based prediction methods, researchers have used co-evolutionary signals derived from large genomic datasets produced by high-throughput sequencing. Using this data, they perform multiple sequence alignment (MSA), i.e., aligning the sequences of a protein of interest (POI) across multiple species to build its comprehensive evolutionary history. Analysis of the MSA reveals two critical patterns for a POI: residues that have been conserved throughout evolution, and pairs of residues that exhibit strong covariance. Such covariance indicates that these residues have changed together over time to maintain productive interactions and follow functional, structural, and folding constraints~\cite{chakrabarti2010structural, sandler2014functional}. Co-evolutionary information is often represented as a contact map, which allows for identification of long-range interactions within protein sequences~\cite{balakrishnan2011learning,morcos2011direct,gobel1994correlated, shindyalov1994can, jones2012psicov,martin2005using}.

MetalNet \cite{cheng2023co} is the first \textit{co-evolution-based} method for identifying metal-binding residue pairs. It focuses on CHED residues (cysteine, histidine, glutamic acid, and aspartic acid), which frequently occur at metal-binding sites.
The method uses the MSA Transformer \cite{pmlr-v139-rao21a} to generate contact maps, where co-evolved CHED pairs are extracted. These pairs are then processed using machine learning (ML) models to predict metal-binding interactions. Finally, MetalNet assembles the predicted pairs into networks and applies motif-matching to identify the metal type.
MetalNet2~\cite{metalnet2} extends MetalNet by incorporating residue-level embeddings from a protein language model (PLM) and replacing motif-matching with an ML-based metal-type predictor. 
Both MetalNet and MetalNet2 demonstrate the utility of co-evolutionary information, but still have limitations. Both methods fail to capture the complete structure of the co-evolved residue network when identifying metal-binding sites. MetalNet relies on motif-matching, which limits its ability to generalize to new network structures or singleton residue pairs, whereas MetalNet2 does not use the network structure and analyses residue pairs in isolation.

To address these limitations, we introduce the Metal-Binding Graph Neural Network (MBGNN), a new co-evolution-based method for predicting metal-binding residues and their corresponding metal types with a focus on CHED residues. Unlike previous methods that analyze residue pairs in isolation or rely on fixed motifs, MBGNN connects co-evolved CHED residue pairs through shared residues to construct co-evolved residue networks. These networks capture the overall structural relationships among residues. 
We apply graph neural networks (GNNs) \cite{khoshraftar2024survey, velivckovic2023everything, corso2024graph} to learn effective residue representations from these networks, enabling the model to capture complex co-evolutionary dependencies and thereby improve metal-binding residue identification and metal-type classification beyond traditional motif-matching methods. To further enhance performance, we implement an $M$-fold ensemble strategy that makes the model less sensitive to class imbalance between binding and non-binding residues, as well as across different metal types. Experimental results demonstrate that MBGNN substantially outperforms existing co-evolution-based and sequence-based baselines, particularly in the precision of metal-binding residue and metal-type prediction.

The remainder of this paper is organized as follows.
Section~\ref{sec:method} presents the MBGNN pipeline, including the construction of co-evolved residue networks, the metal-binding and metal-type predictors, and the $M$-fold ensemble strategy.
Section~\ref{sec:experiments} describes the datasets and the implementation details, introduces the baseline methods, and reports comparative performance results, sensitivity analysis, and ablation studies. Finally, Section~\ref{sec:conclusion} summarizes our findings and outlines future research directions.

\section{Methodology}
\label{sec:method}
MBGNN's pipeline consists of three main components: Co-evolved residue network construction, metal-binding residue identification, and metal-type classification. In this section, we introduce each component and the ensemble strategy. Figure \ref{fig:visualabstract} shows an overview of MBGNN's pipeline.

\subsection{Definition of Co-Evolved Residue Networks}
\label{subsubsec:def}
Given a protein chain with residues indexed by $\{1,\dots,L\}$, let
$\mathrm{score}(i,j)\in\mathbb{R}$ denote a quantitative measure of coordinated variation between residues $i$ and $j$, such as a statistical or learned signal reflecting co-evolution. 
A \emph{co-evolved residue network} is defined as an undirected graph,
\begin{equation}
    G=(\mathcal{V},\mathcal{E},\mathbf{X}),
\end{equation}
where each node $v\in\mathcal{V}$ represents a residue at position $r(v)\in\{1,\dots,L\}$, optionally restricted to a candidate subset. An edge $(u,v)\in\mathcal{E}$ encodes a putative co-evolution relationship, included when the co-evolution value $\mathrm{score}(r(u),r(v))$ satisfies a specified selection criterion (e.g., exceeds a threshold). Edges are undirected. 
Node features $\mathbf{X}\in\mathbb{R}^{|\mathcal{V}|\times d}$ denote initial residue descriptors of dimension $d$.

For a given protein chain, we consider the connected components induced by $\mathcal{E}$ as a collection of subgraphs,
\begin{equation}
\mathcal{G}=\{G_1,G_2,\dots,G_k\},
\end{equation}
where $\mathcal{G}$ is the set of $k$ connected components (subgraphs $G_i$) that capture the co-evolution relationships in the protein chain.

\subsection{Co-Evolved Residue Network Construction}
\label{subsubsec:feat}
To construct the co-evolved residue networks $\mathcal{G}$ for a given protein of interest (POI), we instantiate the graph components through the following procedure.

To generate co-evolutionary scores, we first perform MSA using ColabFold MMseq2 \cite{mirdita2022colabfold}. For compatibility with the MSA Transformer model, we subsample up to $N_{seq}$ aligned sequences by maximizing pairwise Hamming distances. These sequences are then processed by the ESM-MSA-1b model to produce a contact map representing residue-residue co-evolution scores.

The graph components are then defined as follows:
\begin{itemize}
    \item The nodes $\mathcal{V}$ are the CHED residues of the POI's chain.
    \item An edge $(u,v)\in\mathcal{E}$ is added when the contact map score satisfies $\mathrm{score}(r(u), r(v)) > 0.1$.
    \item Each node feature in $\mathbf{X}$ is a $1280$-dimensional feature vector generated by the ESM2 model (esm2-t33-650M-UR50D)~\cite{lin2023evolutionary}. Residue-level embeddings for the full POI sequence are extracted using the PLM-based approach adopted in recent studies~\cite{metalnet2,lmetalsite,mionic, jha2022prediction,jha2023graph, bepler2021learning}.
\end{itemize}

Finally, the graph induced by these nodes and edges is decomposed into connected components, yielding a set of subgraphs $\mathcal{G} = \{G_1, G_2, \dots, G_k\}$. As $\mathcal{G}$ contains all co-evolved residues, it represents the complete co-evolved residue network and preserves the entire co-evolutionary information that existing co-evolution-based methods may fail to capture.

\subsection{Metal-Binding Predictor}
\label{subsec:mbp}
To model the complete co-evolved residue network and classify residues as metal-binding or non-binding, we develop a SAGEConv-based GNN~\cite{hamilton2017inductive} consisting of $N_{bind}$ layers.
SAGEConv is selected for its strong inductive generalization capabilities and its effective use of node features through neighborhood aggregation. By separately aggregating each node’s own representation and the averaged representation of its neighbors, SAGEConv enables the model to retain node-specific features while capturing higher-order dependencies across the complete co-evolved residue network. Each layer iteratively refines node representations by integrating information from both local and global network contexts.

As described in Section~\ref{subsubsec:feat}, each node is initialized with a 1280-dimensional feature vector.
The first layer maps these features into a lower-dimensional hidden representation, and subsequent layers iteratively refine the representations to capture information from more distant nodes in the network.

The SAGEConv operation is defined as:
\begin{equation}
    \mathbf{h}_i' = \mathbf{W}_1 \mathbf{h}_i + \mathbf{W}_2 \cdot \text{mean}_{j \in \mathcal{N}(i)} \mathbf{h}_j,
\end{equation}
where 
\begin{itemize}
    \item \(\mathbf{h}_i'\)  is the updated representation of node \(i\);
    \item \(\mathbf{h}_i\) is the current representation of node \(i\);  
    \item \(\mathcal{N}(i)\) represents the set of neighbors of node \(i\);
    \item \(\mathbf{W}_1\) and \(\mathbf{W}_2\) are learnable weight matrices for self-features and neighbor aggregation, respectively;
    \item \(\text{mean}_{j \in \mathcal{N}(i)} \mathbf{h}_j\) is the mean of neighbor feature vectors.  
\end{itemize}

In our model, each SAGEConv layer, except for the final layer, is followed by batch normalization to stabilize training and a ReLU activation function. The final SAGEConv layer outputs node representations that are passed through a softmax activation function to produce class probabilities for the metal-binding and non-binding classes.

Finally, a new set of graphs $\mathcal{G}'$ is constructed for the metal-type predictor. These graphs are the connected components of the subgraph induced by the residues predicted as metal-binding and follow the general definition in Section~\ref{subsubsec:def}. Edges and node features are inherited from the original networks in $\mathcal{G}$. The resulting graphs $\mathcal{G}'$ are then passed to the metal-type predictor to predict the associated metal type for each residue.

\subsection{Metal-Type Predictor}

The metal-type predictor assigns a specific metal type to each residue in $\mathcal{G}'$.
We again use a SAGEConv-based GNN with $N_{type}$ layers for this prediction task, leveraging its inductive nature and ability to capture local residue features via co-evolutionary relationships.
Each SAGEConv layer, except for the last, is followed by batch normalization and a ReLU activation function. The first SAGEConv layer projects the input features into a lower-dimensional hidden space, and the subsequent layers iteratively refine these representations. The final layer maps the learned residue-level representations to the number of metal types, followed by a softmax activation function that generates the classification probabilities for each metal type.

\subsection{M-Fold Ensemble Strategy}
\label{subsec:mfold}
Due to biological constraints, metal-binding residues are much rarer than non-binding ones, and the distribution of metal types is highly uneven. This class imbalance, common in protein-metal interaction datasets, can bias models toward majority classes and cause overfitting or unstable convergence during training. 
To make the model less sensitive to class imbalance, we adopt an $M$-fold ensemble strategy inspired by Ref \cite{XIA2023168091}. The training set is divided into $M$ folds, and in each iteration, one fold is used for validation while the remaining $(M-1)$ folds are used for training. This process yields $M$ predictors, whose class probability outputs are averaged to obtain the final prediction. This strategy is applied independently to both the metal-binding and metal-type predictors.

\section{Experiments}
\label{sec:experiments} 

In this section, we introduce the dataset and data preprocessing methods used for training and evaluating MBGNN, outline the implementation details and experimental setup, compare MBGNN with other metal-binding site prediction models, and discuss the results. We also provide a sensitivity analysis followed by a qualitative UMAP visualization and ablation studies on the effects of different architectures and PLM-derived embeddings.

\subsection{Dataset and Data Preprocessing}

This subsection describes the data resources. We begin by describing the subsets provided by MetalNet2 and how co-evolution graphs and residue labels are derived for both metal-binding detection and metal-type classification.  We then introduce the dataset used in MIonSite\cite{QIAO201975} for validation of generalization performance.

\subsubsection{MetalNet2 Dataset} 
\label{subsec:dataset}
\begin{figure}[!t]
    \centering
    \includegraphics[width=0.8\columnwidth]{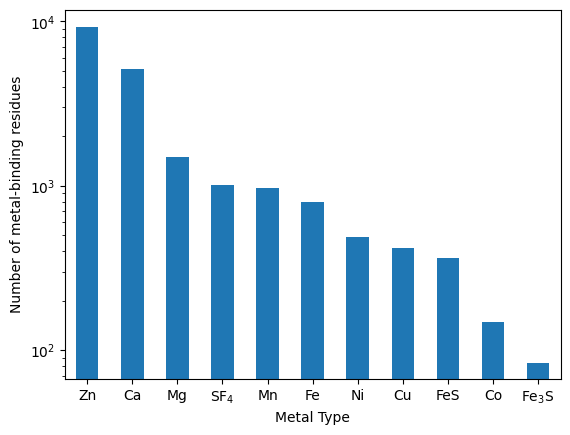}
    \caption{Distribution of metal types in the MetalNet2 dataset.}
    \label{fig:metal_dist}
    \vspace{-10pt}
\end{figure}

The dataset provided by MetalNet2, contains 4,449 metal-binding protein chains collected from the Protein Data Bank (PDB) as of May 2023. The dataset contains a training set and a fixed hold-out test set, which include 18,230 and 1,981 metal-binding CHED residues, respectively. Furthermore, 11 metal types are included for metal-type prediction, which are Zn, Ca, Mg, Mn, Fe, SF$_4$, Ni, Cu, Co, FeS, and Fe$_3$S. Figure~\ref{fig:metal_dist} shows the distribution of metal types, highlighting the class imbalance among them.

Following the procedures explained in Section \ref{subsubsec:feat}, we extracted the graphs from the sequences in the training and test sets, which were used for training and evaluating MBGNN's metal-binding predictor. Metal-binding co-evolved residues in the dataset were labeled as 1, and all the non-binding co-evolved residues were labeled as 0.  Moreover, to prepare the training data for MBGNN's metal-type predictor, training graphs were created by assembling co-evolved metal-binding residues from the training set. For evaluation, we constructed graphs with the residues from the test set that were predicted to be metal-binding for MBGNN's metal-type predictor. Each residue was labeled according to its associated metal type. Table~\ref{tab:bin_graphs} summarizes the number of graphs and co-evolved residues extracted from the sequences for each predictor.

\begin{table}[tb]
\centering
\caption{Summary of Extracted Co-evolved Residues and Graphs for MBGNN's Metal-Binding and Metal-Type Predictors.}
\renewcommand{\arraystretch}{1.3} %
\begin{tabular}{l r}
\toprule\toprule
Metal-Binding Predictor Data & Count \\ 
\midrule
\rowcolor{mygray}
Train Graphs & 38,010 \\ 
Train Co-evolved Residues & 148,163 \\ 
\rowcolor{mygray}
Train Co-evolved Metal-Binding Residues & 17,004 \\
Test Graphs & 4,235 \\
\rowcolor{mygray}
Test Co-evolved Residues & 16,447 \\
Test Co-evolved Metal-Binding Residues & 1,826 \\
\midrule
Metal-Type Predictor Data &  \\ 
\midrule
\rowcolor{mygray}
Train Graphs & 4,153 \\
Test Graphs & 380 ± 7$^{*}$ \\
\bottomrule\bottomrule
\end{tabular}
\label{tab:bin_graphs}

\vspace{0.1cm}
{\raggedright \footnotesize $^{*}$ The number of test graphs for the metal-type predictor depends on the metal-binding predictor's performance, averaged over five runs.\par}
\vspace{-10pt}
\end{table}

\subsubsection{MIonSite Dataset}
\label{subsubsec:indep_benchmark}

To evaluate the generalization performance of MBGNN, we utilized the independent dataset provided by MIonSite~\cite{QIAO201975}. We selected protein chains that were not included in the MetalNet2 dataset and used its CHED residue ground-truth annotations for Zn, Mg, Ca, and Mn metal types (84 protein chains). This allowed us to benchmark MBGNN with other methods and evaluate their generalization performance in metal-type prediction.

\subsection{Implementation Details}
To prepare the input data, we subsampled a maximum of $N_{seq}=64$ aligned sequences after performing MSA with ColabFold MMseq2 to prepare the MSA Transformer input as explained in Section~\ref{subsubsec:feat}.

For training MBGNN’s metal-binding predictor, we followed the $M$-fold ensemble strategy described in Section~\ref{subsec:mfold}. We randomly shuffled and split the extracted metal-binding/non-binding training graphs into six folds ($M=6$),
resulting in six independently trained metal-binding predictor models, with their averaged outputs used for metal-type prediction.
These models were implemented using PyTorch (version 2.3.0)~\cite{10.5555/3454287.3455008} and PyTorch Geometric (version 2.6.1)~\cite{fey2019fast}. The best-performing model was selected based on the F1 score on the validation set. The training was performed for 50 epochs using the Adam optimizer\cite{kingma2014adam} with a learning rate of 0.001, weight decay of 0.0001, and cross-entropy loss as the objective function. The hidden-dimension parameters of all metal-binding predictor models were set to 64, and five SAGEConv layers ($N_{bind}=5$) were used in the model.

For training MBGNN’s metal-type predictor, we applied the same $M$-fold ensemble strategy, splitting the dataset into six folds ($M=6$). The best-performing model for each fold was selected using the F1 score during validation. The final prediction was the average of the outputs of the six selected models. The training hyperparameters remained consistent with those of the metal-binding predictor, except for a weight decay of 0.0005 in the Adam optimizer and the hidden dimension size of 512. Metal-type predictor models were trained for 500 epochs. This model also used five SAGEConv layers ($N_{type}=5$).

Hyperparameter selection and optimization for both metal-binding predictor and metal-type predictor were performed using Optuna (version 4.1.0) \cite{akiba2019optuna}. For the tuning process, a single representative model was selected. This model was trained on its designated training folds and evaluated on its corresponding validation fold to guide the optimization. The hyperparameters and their respective search spaces explored during this process are detailed in Table \ref{tab:hyperparameters}.

\begin{table}[tb!]
\centering
\caption{Hyperparameter Search Space for Model Optimization}
\label{tab:hyperparameters}
\begin{tabular}{ll}
\toprule\toprule
Hyperparameter & Search Space \\ 
\midrule
\rowcolor{mygray}
Number of Layers & $\{1, 2, 3, 4, 5, 6\}$ \\
Learning Rate & $\{0.001, 0.01, 0.05, 0.1\}$ \\
\rowcolor{mygray}
Epochs & Integer from $[50, 1000]$ with step 50 \\
Hidden Dimension & Integer from $[16, 512]$ with step 16 \\
\rowcolor{mygray}
Weight Decay & Uniform float from $[1 \times 10^{-6}, 1 \times 10^{-3}]$ \\ 
\bottomrule\bottomrule
\vspace{-10pt}
\end{tabular}
\end{table}

\begin{table}[tb]
\centering
\caption{Performance Comparison for Metal-Binding and Metal-Type Predictions on the MetalNet2 test set (\textbf{Best}, \underline{Second Best}).}
\label{tab:performance_comparison}

\begingroup
\setlength{\tabcolsep}{3pt} %

\begin{tabular}{lcccc} %
\toprule\toprule
Metric        & MetalNet\cite{cheng2023co} & MetalNet2\cite{metalnet2} & M\text{-}Ionic\cite{mionic} & MBGNN (Ours) \\ 
\midrule
\multicolumn{5}{c}{Metal-Binding Prediction} \\ 
\midrule
\rowcolor{mygray}
Precision & 0.666 & 0.826 & \underline{0.880} & \textbf{0.882} \\
Recall    & 0.566 & \underline{0.700} & 0.557 & \textbf{0.703} \\
\rowcolor{mygray}
F1 Score  & 0.612 & \underline{0.758} & 0.682 & \textbf{0.783} \\ 
\midrule
\multicolumn{5}{c}{Metal-Type Prediction} \\ 
\midrule
\rowcolor{mygray}
Precision & 0.153 & \underline{0.525} & 0.315 & \textbf{0.719} \\
Recall    & 0.201 & \textbf{0.535} & 0.269 & \underline{0.492} \\
\rowcolor{mygray}
F1 Score  & 0.108 & \underline{0.521} & 0.285 & \textbf{0.554} \\ 
\bottomrule\bottomrule
\end{tabular}
\vspace{-10pt}
\endgroup
\end{table}

\subsection{Baseline Methods}
\label{subsec:baselines}

In this paper, we compare our method with other advanced co-evolution-based and sequence-based methods for metal-binding site prediction, which we briefly review below.

\begin{itemize}

    \item \textbf{MetalNet}\cite{cheng2023co} is the first co-evolution-based method for predicting metal-binding sites. It performs MSA and uses the MSA Transformer to generate contact maps, from which co-evolved CHED residue pairs are extracted. A classifier is then trained using pairwise amino-acid frequency matrices derived from the MSA to label these CHED pairs as metal-binding or not. The predicted pairs are assembled into co-evolution networks, and metal types are assigned by matching these networks against a curated fixed motif bank. For a fair comparison, we re-implemented MetalNet based on its original paper and source code. The model was trained using the MetalNet2 training set and the motif bank extracted from the MetalNet2 dataset. All hyperparameters were set the same as in the original paper.

    \item \textbf{MetalNet2}\cite{metalnet2} extends MetalNet and achieves superior performance by incorporating a larger training set and using pair encodings derived from ESM2 to improve feature representation. 
    We used the same training dataset as in the original paper and their provided pre-trained model weights from the official repository in the comparison.

    \item \textbf{LMetalSite}\cite{lmetalsite} is an alignment-free, sequence-based predictor for four metal types (Zn, Ca, Mg, Mn), which is an upstream model of the field, used for comparison in recent years. It uses pretrained protein language-model embeddings, a Transformer encoder, and multi-task learning across ions. The results reported in this study were obtained via its publicly available web server\footnote{LMetalSite: {\scriptsize\url{https://bio-web1.nscc-gz.cn/app/lmetalsite}}}, as the authors did not release a training script.
    
    \item \textbf{M-Ionic}\cite{mionic} is a recent state-of-the-art sequence-based method that uses residue-level embeddings from pretrained protein language models as input to a Transformer-based classifier to identify metal-binding proteins and residues across multiple metal types. For a fair comparison, we retrained M-Ionic on the MetalNet2 training set using the same hyperparameters reported in the original paper.
    
\end{itemize}

\subsection{Results}
\label{subsec:res}
We first report results on the MetalNet2 dataset in Table~\ref{tab:performance_comparison} and Fig.~\ref{fig:allres}, followed by the MIonSite dataset results in Fig.~\ref{fig:mion_bench}.

\subsubsection{Results on MetalNet2 Dataset}

\begin{figure*}[t]
    \centering
    \subfloat[]{
        \includegraphics[width=0.95\linewidth]{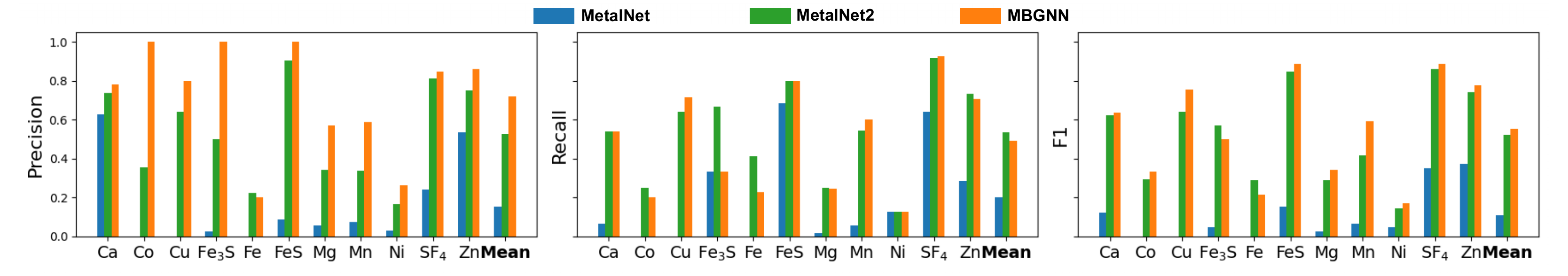}
        \label{fig:a}}
        \hfil
    \subfloat[]{
        \includegraphics[width=0.95\linewidth]{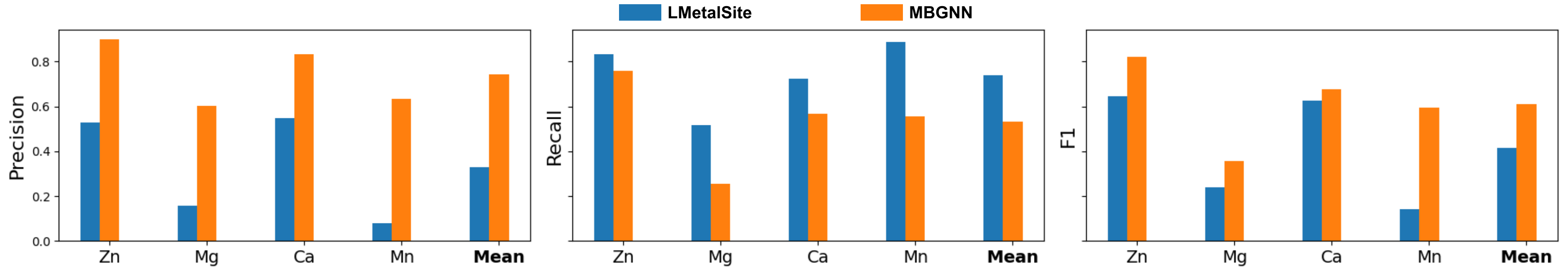}
        \label{fig:b}}
        \hfil
    \subfloat[]{
        \includegraphics[width=0.95\linewidth]{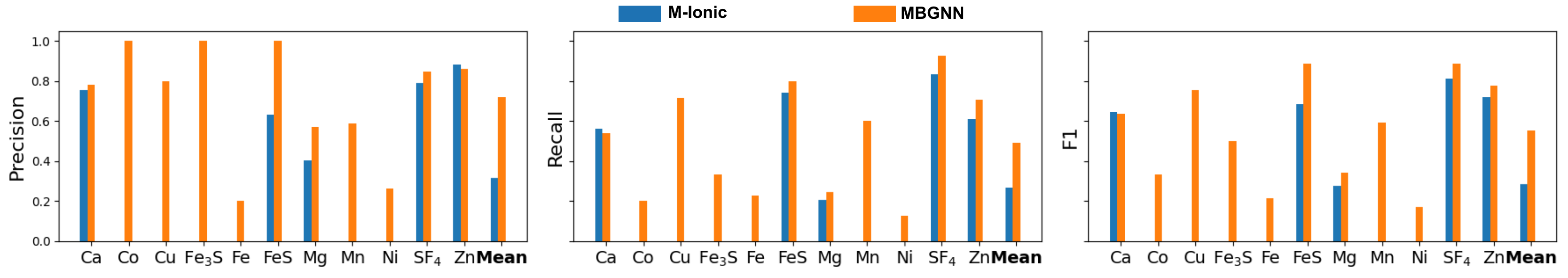}
        \label{fig:c}}
    \caption{Performance comparison of MBGNN for metal-type prediction on the MetalNet2 test set with (a) co-evolution-based methods and sequence-based methods, (b) LMetalSite~\cite{lmetalsite} and (c) M-Ionic~\cite{mionic}. In all subfigures, the ``Mean'' column represents the macro average of the values of the metal predictions.
    }
    \label{fig:allres}
\vspace{-10pt}
\end{figure*}

\paragraph{Comparison with co-evolution-based baselines}
We compared the performance of MBGNN with two co-evolution-based methods, MetalNet and MetalNet2. We reported the precision, recall, and F1 scores for both metal-binding residue and metal-type predictions on the MetalNet2 test set. Table~\ref{tab:performance_comparison} presents the performance for metal-binding residue prediction and the macro average of metrics for metal-type prediction, while Fig.~\ref{fig:a} shows the performance for the 11 metal types. 

As shown in Table~\ref{tab:performance_comparison}, compared to MetalNet, MBGNN significantly improves all evaluation metrics in both metal-binding and metal-type prediction, demonstrating superior overall predictive performance.
Compared with MetalNet2, MBGNN achieved higher precision in both metal-binding and metal-type prediction by 5.6 and 19.4 percentage points, respectively, demonstrating its ability to reduce false positives while maintaining strong recall.
For the overall F1 score, MBGNN shows absolute improvements of 2.5 and 3.3 percentage points on the two tasks, respectively, showing better balance between precision and recall.

As shown in Fig.~\ref{fig:a}, these improvements are also evident for metals with limited training samples, such as Co, Ni, and FeS, where MBGNN outperformed other methods. 
We observed that MetalNet2 achieved higher recall for some metal types by predicting multiple metals per residue due to its pairwise formulation, which treated residue pairs independently. This approach artificially increased recall but reduced precision, often assigning multiple metal types to a single residue and generating more false positives. In contrast, MBGNN mitigates this issue by leveraging the complete network of co-evolved residue interactions through GNNs, enabling more precise predictions while maintaining strong recall. Consequently, MBGNN achieves higher precision, F1 scores, and overall reliability in both tasks.

\paragraph{Comparison with sequence-based baselines}
We then compared MBGNN with the two sequence-based approaches introduced in Section~\ref{subsec:baselines}.

We first compared MBGNN with LMetalSite, which predicts only four essential metal types for residues: Zn, Mg, Ca, and Mn. As the training script of LMetalSite is unavailable, we ensured a fair comparison by extracting from the hold-out test set only residues corresponding to these four metal types, excluding protein chains present in LMetalSite’s training data. We then evaluated the classification performance of both models on these chains' CHED residues. The results shown in Fig.~\ref{fig:b} demonstrate that MBGNN significantly outperformed LMetalSite in precision and F1 score across all four metal types and the average. These findings highlighted MBGNN’s ability to make accurate predictions by effectively leveraging co-evolutionary information. 
Specifically, the higher precision of MBGNN highlights its ability to reduce false positives while maintaining strong predictive performance. In contrast, LMetalSite achieved higher recall, likely because it considers residues without co-evolutionary relationships, allowing it to classify a broader range of residues. These results suggest that although co-evolutionary information is highly valuable, integrating it with complementary sequence- or structure-based features could further enhance prediction coverage.

We then compared MBGNN with the M-Ionic model on the MetalNet2 test set, with results summarized in Table~\ref{tab:performance_comparison}. MBGNN demonstrates a clear advantage, achieving an F1 score of 0.783 for metal-binding (compared to 0.682 for M-Ionic) and 0.554 for metal-type prediction (compared to 0.285). Moreover, as shown in Fig.~\ref{fig:c}, MBGNN achieved higher precision, recall, and F1 scores for most individual metals as well as in the macro average.
The precision gain was especially pronounced for underrepresented metals such as Co and FeS, highlighting the benefit of exploiting the complete co-evolved residue network. The macro-average F1 score across all 11 metals further confirms the robustness of MBGNN and shows that integrating co-evolutionary context leads to more balanced predictions than analyzing residues in isolation.
This improvement is especially meaningful because both methods use the same ESM2 embeddings as initial residue features.

\subsubsection{Results on MIonSite Dataset}

\begin{figure*}[tbh]
  \centering  \includegraphics[width=0.9\linewidth]{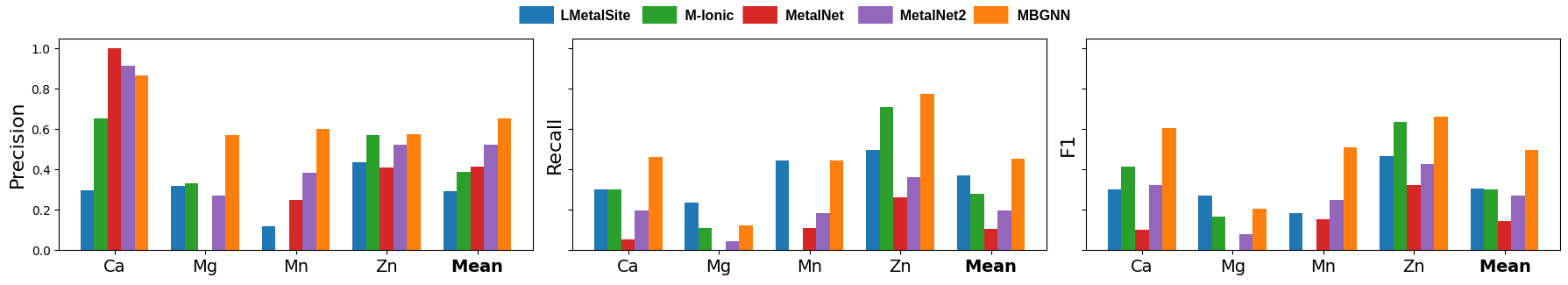}
  \caption{Performance comparison of MBGNN for metal-type prediction on the MIonSite test set. 
  }
  \label{fig:mion_bench}
  \vspace{-10pt}
\end{figure*}

We evaluated all models on the non-redundant MIonSite benchmark to assess model generalizability. Fig.~\ref{fig:mion_bench} shows the same performance ranking observed on the MetalNet2 test set. Across all four metal types, MBGNN achieved the best balance between precision and recall, leading to the highest F1 scores. LMetalSite and MetalNet2 followed, each showing strengths only in isolated metrics for specific metals. The consistency of these results across an independent dataset underscores the effectiveness of modeling the complete network of co-evolved residues rather than analyzing residues in isolation. This alignment between two independent datasets further demonstrates the robustness and broad applicability of the co-evolutionary graph representation employed by MBGNN.

\subsection{Model Stability across Random Seeds}
\label{subsec:sens}
To assess model robustness, we repeated the entire training and testing process of the pipeline five times, each with a different random seed. The results, summarized in Table~\ref{tab:mbgnn_results}, demonstrate the stability and consistency of the model’s performance across runs. This stability is evidenced by the low standard deviations (Std) reported in the final column of Table~\ref{tab:mbgnn_results}. For instance, the F1 Score for Metal-Binding Prediction was $0.783 \pm 0.004$ and for Metal-Type Prediction was $0.551 \pm 0.003$, indicating high consistency across different initializations. The best F1 score from the metal-type classifier, obtained in the fifth run, was selected for the final predictor.

\begin{table}[tb]
\centering
\caption{Performance of MBGNN across Five Training Runs (1-5) on the MetalNet test set, Reported as Mean $\pm$ Std, where Std is the Standard Deviation.}
\label{tab:mbgnn_results}
\begin{tabular}{l|ccccc|c}
\toprule\toprule
Metric & 1 & 2 & 3 & 4 & 5 & Mean $\pm$ Std \\ \midrule
\multicolumn{7}{c}{Metal-Binding Prediction} \\ 
\midrule
\rowcolor{mygray}
Precision  & 0.888 & 0.875 & 0.877 & 0.881 & 0.882 & 0.881 $\pm$ 0.005 \\
Recall     & 0.702 & 0.715 & 0.710 & 0.695 & 0.703 & 0.705 $\pm$ 0.008 \\
\rowcolor{mygray}
F1 Score   & 0.784 & 0.787 & 0.785 & 0.777 & 0.783 & 0.783 $\pm$ 0.004 \\ 
\midrule
\multicolumn{7}{c}{Metal-Type Prediction} \\ 
\midrule
\rowcolor{mygray}
Precision  & 0.704 & 0.689 & 0.711 & 0.675 & 0.719 & 0.700 $\pm$ 0.018 \\
Recall     & 0.493 & 0.495 & 0.493 & 0.497 & 0.492 & 0.494 $\pm$ 0.002 \\
\rowcolor{mygray}
F1 Score   & 0.547 & 0.548 & 0.553 & 0.552 & 0.554 & 0.551 $\pm$ 0.003 \\ 
\bottomrule\bottomrule
\end{tabular}
\vspace{-10pt}
\end{table}

\subsection{UMAP Visualization of Raw and Final Hidden Embeddings}

\begin{figure*}[t]
  \centering
\includegraphics[width=0.8\textwidth]{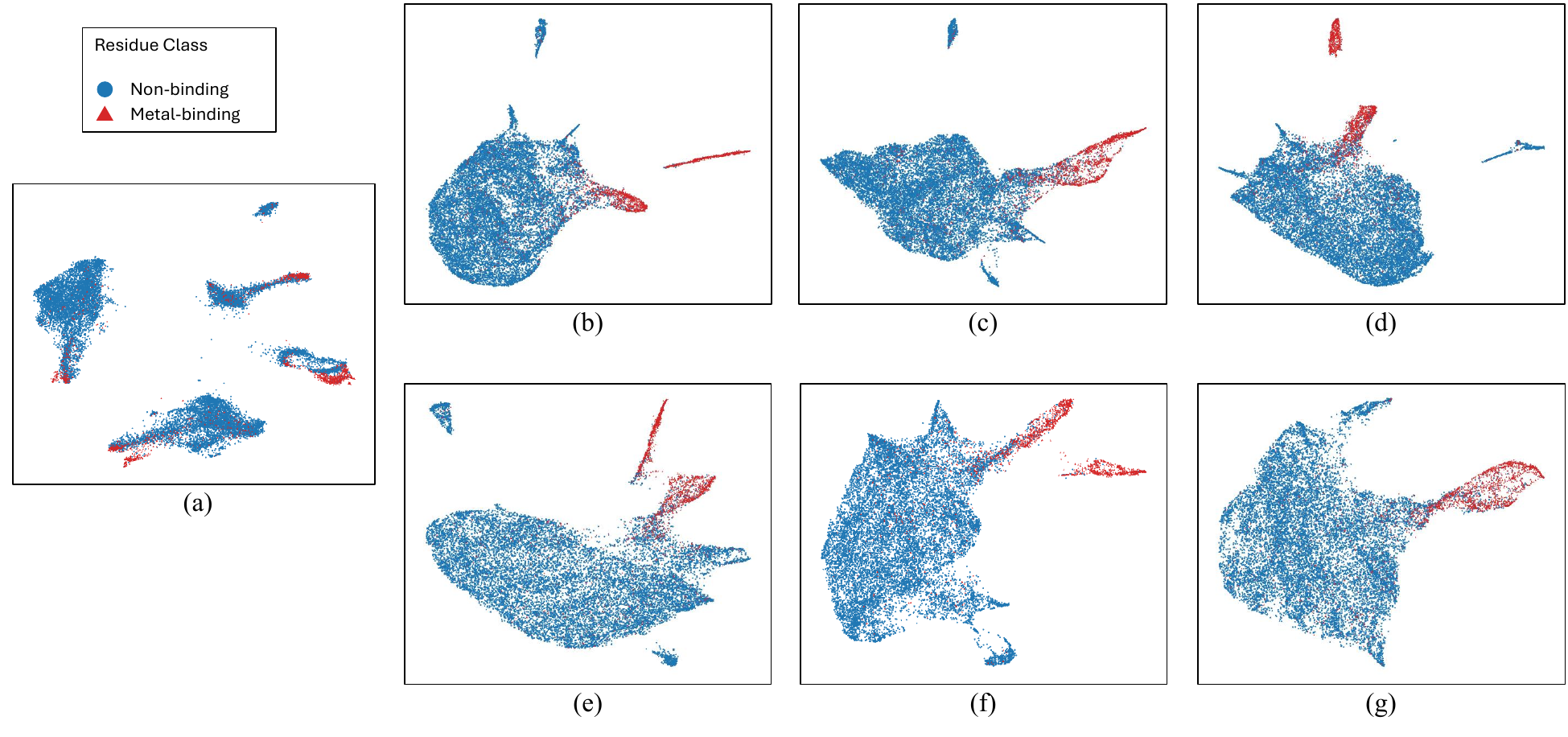}
  \caption{UMAP projections on the MetalNet2 test set. 
  (a-g) Residue embeddings colored by metal-binding versus non-binding: 
  (a) raw ESM2 embeddings; (b-g) MBGNN-learned embeddings from each model in MBGNN's metal-binding predictor ensemble.}
  \label{fig:umap_binding}
  \vspace{-5pt} 
\end{figure*}

\begin{figure*}[ht]
  \centering
\includegraphics[width=0.8\textwidth]{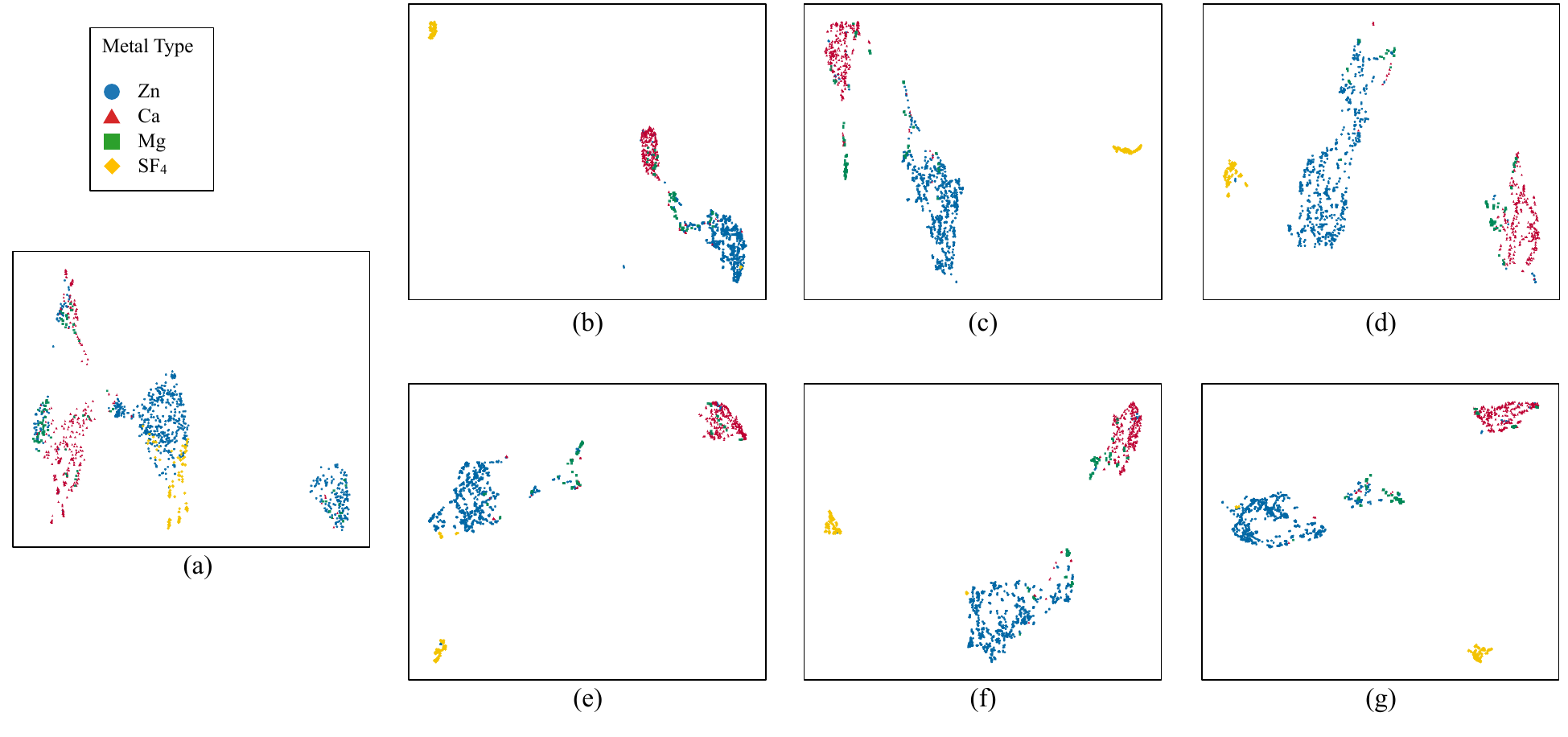}
  \caption{UMAP projections on the MetalNet2 test set. 
  (a-g) Embeddings for metal-binding residues coloured by metal type: 
  (a) raw ESM2 embeddings; (b-g) MBGNN-learned embeddings from each model in MBGNN's metal-type predictor ensemble.}
  \label{fig:umap_metaltype}
  \vspace{-10pt}
\end{figure*}

To visualize how MBGNN transforms residue features, Figures~\ref{fig:umap_binding} and~\ref{fig:umap_metaltype} compare raw and learned embeddings using UMAP~\cite{2018arXivUMAP}. For metal-binding versus non-binding, Fig.~\ref{fig:umap_binding}a shows the \emph{raw ESM2} per-residue embeddings, while Fig.~\ref{fig:umap_binding}b-g show the \emph{final hidden} embeddings generated by the last SAGEConv layer before classification. For metal type, Fig.~\ref{fig:umap_metaltype}a visualizes the raw ESM2 embeddings for metal-binding residues, and Fig.~\ref{fig:umap_metaltype}b-g display the corresponding MBGNN-learned embeddings. For clarity, the metal-type visualizations include only the four most frequent types, $\mathrm{Zn}$, $\mathrm{Ca}$, $\mathrm{Mg}$, and $\mathrm{SF_4}$, which provide sufficient samples for a reliable projection.

Compared with the raw ESM2 features, the final hidden embeddings form tighter and more distinct clusters, particularly across metal types. This indicates that MBGNN effectively incorporates co-evolutionary neighborhood information to sharpen structural distinctions between classes and metal types. Here, UMAP serves solely as a neighborhood-preserving tool for qualitative visualization.

\subsection{Ablation Studies}
\label{subsec:ablation}
To evaluate the impact of (i) the GNN layer type and (ii) the choice of residue-level PLM embeddings, we conducted two controlled ablation experiments on the MetalNet2 test set. In each experiment, we changed \emph{one} component while keeping all other components, hyperparameters, and the $M$-fold ensemble protocol unchanged.

\begin{table}[t]
\centering
\caption{Ablation studies of different GNN layers on the MetalNet2 test set (\textbf{Best}, \underline{Second Best}).
}
\label{tab:ablation-gnn-layers}
\begin{tabular}{lccc}
\toprule\toprule
Metric &  GCNConv~\cite{kipf2016semi} & GATV2Conv~\cite{brody2021attentive} & SAGEConv \\
\midrule
\multicolumn{4}{c}{Metal-Binding Prediction} \\ 
\midrule
\rowcolor{mygray}
Precision & 0.780 & \underline{0.832} & \textbf{0.881} \\
Recall & \underline{0.656} & 0.648 & \textbf{0.705} \\
\rowcolor{mygray}
F1 Score & 0.712 & \underline{0.728} & \textbf{0.783} \\
\midrule
\multicolumn{4}{c}{Metal-Type Prediction} \\ 
\midrule
\rowcolor{mygray}
Precision & 0.531 & \underline{0.607} & \textbf{0.700} \\
Recall & \underline{0.476} & 0.426 & \textbf{0.494} \\
\rowcolor{mygray}
F1 Score & \underline{0.490} & 0.482 & \textbf{0.551} \\
\bottomrule\bottomrule
\end{tabular}
\vspace{-5pt}
\end{table}

\begin{table}[t]
\centering
\caption{Ablation studies of different initial embeddings on the MetalNet2 test set (\textbf{Best}, \underline{Second Best}).
}
\label{tab:ablation-embeddings}
\begin{tabular}{lccc}
\toprule\toprule
Metric &  ProtTrans~\cite{9477085} & ESM C~\cite{esm2024cambrian} & ESM2 \\
\midrule
\multicolumn{4}{c}{Metal-Binding Prediction} \\ 
\midrule
\rowcolor{mygray}
Precision & \underline{0.877} & 0.831 & \textbf{0.881} \\
Recall & \underline{0.662} & 0.610 & \textbf{0.705} \\
\rowcolor{mygray}
F1 Score & \underline{0.754} & 0.703 & \textbf{0.783} \\
\midrule
\multicolumn{4}{c}{Metal-Type Prediction} \\ 
\midrule
\rowcolor{mygray}
Precision & \underline{0.576} & 0.284 & \textbf{0.700} \\
Recall & \underline{0.453} & 0.192 & \textbf{0.494} \\
\rowcolor{mygray}
F1 Score & \underline{0.491} & 0.209 & \textbf{0.551} \\
\bottomrule\bottomrule
\vspace{-10pt}
\end{tabular}
\end{table}

\subsubsection{Different GNN Layers}
\label{sssec:ablation:gnn}
We compared the SAGEConv layers in both the metal-binding and metal-type predictors with GCNConv~\cite{kipf2016semi} or GATV2Conv~\cite{brody2021attentive} layers, while keeping node features the same using the ESM2 embeddings. The results, summarized in Table~\ref{tab:ablation-gnn-layers}, demonstrate the superior performance of SAGEConv. For metal-binding prediction, the baseline model achieved an F1 score of $0.783$, outperforming the GATV2Conv ($0.728$) and GCNConv ($0.712$) variants. A similar pattern was observed for metal-type prediction, where the baseline’s F1 score of $0.551$ exceeded those of GATV2Conv ($0.482$) and GCNConv ($0.490$).

\subsubsection{Different Initial Residue Embeddings}
\label{sssec:ablation:plm}
We kept the original SAGEConv architecture but replaced the ESM2 (esm2-t33-650M-UR50D) embeddings with (i) ProtTrans (ProtT5-XL-UniRef50) embeddings~\cite{9477085} and (ii) ESM~C (esmc-300m-2024-12) embeddings from the EvolutionaryScale ESM~C family~\cite{esm2024cambrian}. The results, summarized in Table~\ref{tab:ablation-embeddings}, demonstrate the advantage of using ESM2 embeddings. ProtTrans embeddings reduced the F1 scores for metal-binding and metal-type prediction to $0.754$ and $0.491$, respectively. ESM~C embeddings led to a more pronounced degradation, particularly for metal-type prediction, where the F1 score dropped to $0.209$.

\section{Discussion}
\label{sec:discussion}

Protein-metal interactions are fundamental to protein structure, catalysis, and regulation, with a substantial fraction of the proteome depending on metal cofactors for proper function. Consequently, the disruption of metal binding is linked to numerous pathologies, highlighting the critical importance of accurately identifying metal-binding residues for biomedical and clinical research.

A major clinical application of metal-binding site prediction is the interpretation of disease-associated genetic variants. Missense mutations that alter metal coordination can reduce enzyme activity, destabilize protein structures, or disrupt signaling pathways. In large-scale sequencing studies, where experimental validation is costly and low-throughput, accurate computational identification of metal-binding sites is essential for prioritizing variants with likely functional and clinical impact. 
To this end, we propose MBGNN, which models the complete network of co-evolved residues to address key limitations of previous methods. As demonstrated in Section \ref{subsec:res}, MBGNN improves both predictive performance and generalization, thereby facilitating hypothesis generation in studies of genetic disorders and complex diseases.

Beyond identifying binding sites, accurately distinguishing metal types is essential for elucidating disease-specific mechanisms. Metal ions play a pivotal role in the tumor microenvironment, where they influence metabolism, oxidative stress, and immune regulation~\cite{liu2025interpretable}. Therefore, metal-dependent proteins and pathways have emerged as clinically relevant components of cancer biology. While MBGNN is not intended for direct clinical diagnosis, its ability to distinguish metal types, including underrepresented ions that are often difficult to characterize computationally, can support the annotation of tumor-related proteins and advance understanding of the metal-dependent mechanisms underlying tumor progression and immune evasion.

In addition, MBGNN shows robust performance on the underrepresented ions, which challenges existing methods due to limited verified annotations and chemically heterogeneous coordination environments~\cite{10.1093/bioinformatics/btac358}. 
Copper (Cu) provides a representative example. Cu-binding proteins are closely linked to disease-relevant redox and metabolic pathways, and copper dysregulation is increasingly implicated in cancer biology and therapeutic strategies~\cite{DASILVA2022111634,chen2022copper}. 
In particular, the mechanism of cuproptosis directly links copper to mitochondrial metabolism, highlighting Cu-dependent vulnerabilities that may be clinically actionable~\cite{tsvetkov2022,chen2022copper}. 
However, Cu-binding residues are intrinsically difficult to predict, because Cu ions adopt diverse, context-dependent coordination geometries that render simple sequence motifs unreliable~\cite{RUBINO2012129}. As shown in Fig~\ref{fig:allres}, MBGNN achieves higher precision and overall F1 scores than recent baselines such as M-Ionic~\cite{mionic}, enabling more accurate Cu-binding residue prediction.

MBGNN also exhibits similar advantages for other clinically important yet underrepresented metal cofactors, such as SF4 ([4Fe-4S]). Fe-S cofactors are essential for mitochondrial respiration and redox regulation and are implicated in a range of disease mechanisms~\cite{READ2021102164}. 
However, their redox and oxygen sensitivity often lead to mis-annotation, with several proteins initially classified as zinc-binding later shown to harbour Fe–S clusters~\cite{10.1093/mtomcs/mfae025,PRITTS2022111756}. 
As shown in Fig.~\ref{fig:umap_metaltype}, MBGNN can clearly separate SF4 sites (yellow diamond) from structurally similar Zn-binding sites (blue circle), indicating that its co-evolutionary graph features capture Fe-S-specific signatures beyond raw embedding similarity. This capability can improve functional annotation and strengthen the variant interpretation in pathways dependent on Fe-S proteins, including mitochondrial metabolism and DNA replication and repair~\cite{shi2021biogenesis}.

Furthermore, accurate prediction of metal-binding sites and metal types is highly relevant to drug discovery and target analysis. Many drug targets are metal-dependent enzymes or regulatory proteins whose functional or active sites are defined by coordinated metal ions. Additionally, metal-based drugs, such as platinum compounds, interact extensively with cellular proteins, contributing to both therapeutic efficacy and toxicity. Accurate characterization of metal-binding sites can therefore support early-stage target evaluation and improve the interpretation of off-target interactions and resistance mechanisms.

Finally, a practical advantage of MBGNN is its reliance on sequence-derived evolutionary information rather than experimentally determined protein structures. Because many clinically relevant proteins lack structural data and experimental annotation of metal-binding sites remains limited, MBGNN enables robust prediction when structural information is unavailable. This makes the method well-suited for the large-scale analysis of disease-associated proteomes, where data availability is often uneven.

\section{Conclusion}
\label{sec:conclusion}
This study presents MBGNN, a novel computational framework for predicting metal-binding residues using co-evolutionary information. MBGNN models the complete network of co-evolved residues using GNNs to identify metal-binding residues and classify them into 11 metal types. By leveraging a GNN-based architecture on co-evolved residue networks, MBGNN addresses key limitations of prior co-evolution methods. MBGNN substantially improves precision and F1 scores for both binding-residue and metal-type prediction, demonstrates strong generalization across datasets, and enhances detection of underrepresented metals. Experimental results and ablation studies demonstrate the effectiveness of MBGNN, compared with four advanced baseline methods. The discussion further highlights the potential applications and impact of MBGNN in clinical research.

A current limitation of MBGNN is its inability to classify residues without co-evolutionary information, which restricts its applicability under limited data. Future work could address this by integrating co-evolutionary networks with structural or sequence-derived features in a multimodal framework.

\section*{References}
\bibliographystyle{IEEEtran}
\bibliography{main}

\end{document}